\documentclass[nohyperref]{article}

\usepackage{microtype}
\usepackage{graphicx}
\usepackage{subfigure}
\usepackage{booktabs} 

\usepackage[hyphens]{url}
\usepackage[pagebackref=true]{hyperref} 
 \renewcommand*\backref[1]{\ifx#1\relax \else (Cited on #1) \fi}

\usepackage[accepted]{icml2022}

\usepackage{amsmath}
\usepackage{amssymb}
\usepackage{mathtools}
\usepackage{amsthm}

\usepackage[capitalize,noabbrev]{cleveref}

\theoremstyle{plain}

\theoremstyle{definition}

\theoremstyle{remark}

\usepackage[textsize=tiny]{todonotes}

\icmltitlerunning{Multilingual Disinformation Detection for Digital Advertising}

\begin{document}

\twocolumn[
\icmltitle{Multilingual Disinformation Detection for Digital Advertising}



\icmlsetsymbol{equal}{*}

\begin{icmlauthorlist}
\icmlauthor{Žofia Trsťanová}{criteo}
\icmlauthor{Nadir El Manouzi}{criteo}
\icmlauthor{Maryline Chen}{criteo}
\icmlauthor{Andre L. V. da Cunha}{criteo}
\icmlauthor{Sergei Ivanov}{criteo}
\end{icmlauthorlist}

\icmlaffiliation{criteo}{Criteo AI Lab, Paris, France}


\icmlcorrespondingauthor{Žofia Trsťanová}{zofia.trstanova@gmail.com}
\icmlcorrespondingauthor{Nadir El Manouzi}{n.elmanouzi@criteo.com}
\icmlcorrespondingauthor{Maryline Chen}{ma.chen@criteo.com}

\icmlkeywords{propaganda detection, disinformation, fake news, ad tech, digital advertisement, embedding}

\vskip 0.3in
]



\printAffiliationsAndNotice{}  

\begin{abstract}
In today's world, the presence of online disinformation and propaganda is more widespread than ever. Independent publishers are funded mostly via digital advertising, which is unfortunately also the case for those publishing disinformation content. The question of how to remove such publishers from advertising inventory has long been ignored, despite the negative impact on the open internet. In this work, we make the first step towards quickly detecting and red-flagging websites that potentially manipulate the public with disinformation. We build a machine learning model based on multilingual text embeddings that first determines whether the page mentions a topic of interest, then estimates the likelihood of the content being malicious, creating a shortlist of publishers that will be reviewed by human experts. Our system empowers internal teams to proactively, rather than defensively, blacklist unsafe content, thus protecting the reputation of the advertisement provider.
\end{abstract}

\section{Introduction}
In recent years, traditional news outlets such as newspapers, magazines, and television are being replaced as main sources of information in favor of social media, podcasts, messaging applications, and online websites  \cite{Pew2022, Pew2021, Pew2018, Reuters2021}. This shift to digital news consumption has been accompanied by an online environment where news pieces from serious institutions increasingly compete for the public's attention against disinformation, fake news, and propagandistic content. As such, the prevalence of untruthful and misleading online content has become a source of concern for governments and other official institutions. Recent events where online disinformation played a crucial role include the COVID19 pandemic \cite{Khan2022, Endo2022} and the 2016 US elections \cite{Zhou2020Theory, Zhou2019, Wang2018, Farajtabar2017}. It also led to a significant economic impact \cite{Rapoza2017}. As companies shift their advertisement budget to online campaigns \cite{Pew2021Sheet}, the preponderance of online disinformation has become a source of concern in the ad tech industry, where advertisers\footnote{An \textbf{advertiser} is a company who pays for the possibility of showing ads for its products on the internet.} and providers do not want to associate themselves with or help fund such publishers\footnote{A \textbf{publisher} is a website that receives revenue by accepting to display ads on its pages.}.

In this work, we propose a pipeline whereby ad providers can detect the presence of undesired content among their inventory and proactively block it. As there are billions of websites, the first step in our solution is to filter the web pages on a topic that can be harmful to the ad providers. We then apply a classifier that predicts the probability of each publisher being malicious and uses it as a ranking score. Finally, trained practitioners inspect the top-ranked publishers to decide if they should be blocklisted or not. Our procedure is intended to assist internal teams in assessing the quality of the publisher's content on a sensitive topic proactively, rather than in a post-hoc manner.

The remainder of this paper is organized as follows: Section \ref{sec:related_work} presents related work; in Section \ref{sec:methods} we state the main problem and describe our methodology and define our solution, the REDD model; in Section \ref{sec:experiments} we perform experiments to address the following questions: Why did we choose fine-tuned embeddings for topic projection? Is the topic projection step necessary to train the disinformation ranking model? What should be used as the input to that model: embeddings or text? How does the multilingual setting impact the performance? In the final section, we highlight the main conclusions and discuss the next steps.

\section{Related Work}
\label{sec:related_work}
Previous literature on disinformation detection focused mostly on social media content. Among existing methods, \textit{propagation-based} and \textit{content-based} are two commonly used techniques. Propagation-based methods analyze the way the content circulates in the social media platform: who produced it, who spread it, and how producers relate to each other \cite{Zhou2020Theory, zhou2018fake, wu2015false, castillo2011information}. Content-based methods attempt to identify disinformation by analyzing its textual and image contents and can generally be grouped as combinatorial \cite{perez2017automatic, shi2016discriminative, ciampaglia2015computational} or neural \cite{Zhou2020Safe, Wang2018}. Combinatorial methods explicitly represent textual aspects known to be indicative of suspicious content and usually rely on extensive feature engineering \cite{Endo2022, Horak2021, Zhou2020Theory, Wang2017Liar}. Neural methods, on the other hand, learn representations with neural networks based on the raw text or image content. Architectures explored in the literature include LSTM, GRU, Bidirectional GRU \cite{Endo2022}, CNN, Bidirectional LSTM \cite{Wang2017Liar}, Text-CNN \cite{Zhou2020Safe, Wang2018}, and BERT \cite{Vorakitphan2022}. Following trends in other Natural Language Processing tasks, neural networks typically outperform the combination of feature engineering and standard classification algorithms, while being easier to train.

Alternatively, some works have focused on detecting propaganda in news articles. \citealt{Vorakitphan2022} employ a BERT-based model for spam identification and use RoBERTa-based model to extract sentence embeddings, which are combined with a series of hand-crafted features \cite{Vorakitphan2021} and fed into a Bi-LSTM model. \citealt{Blaschke2020} adopt a similar approach, combining BERT embeddings and hand-crafted features and feeding them into Bi-LSTM, multi-layer perception, and linear models, but using a more reduced set of features. \citealt{Dao2020} use an LSTM model on top of GloVe embeddings for span identification, and another LSTM model over BERT embeddings for classification. \citealt{SanMartino2020} perform a literature review on computational propaganda detection, confirming that the most performant models are BERT-based. A work similar to ours is that of \citealt{Chang2021}, which selected a list of Twitter accounts with potential propaganda, obtained tweets from them, and manually labeled them according to 21 labels comprising 18 propaganda techniques. The authors then trained a BERT model for multi-class sentence classification. In our work, we adopt a similar, but simpler output: a binary classification, without any further categorization or detection of relevant text spans, and we focus on multi-language publisher content, which is different in nature from user-generated social media content. Also, we use specific embeddings, which were fine-tuned for a multi-label classification task.

\section{Embedding-Based Disinformation Detection}
\label{sec:methods}

\begin{figure}[ht]
\begin{center}
\centerline{\includegraphics[width=\columnwidth]{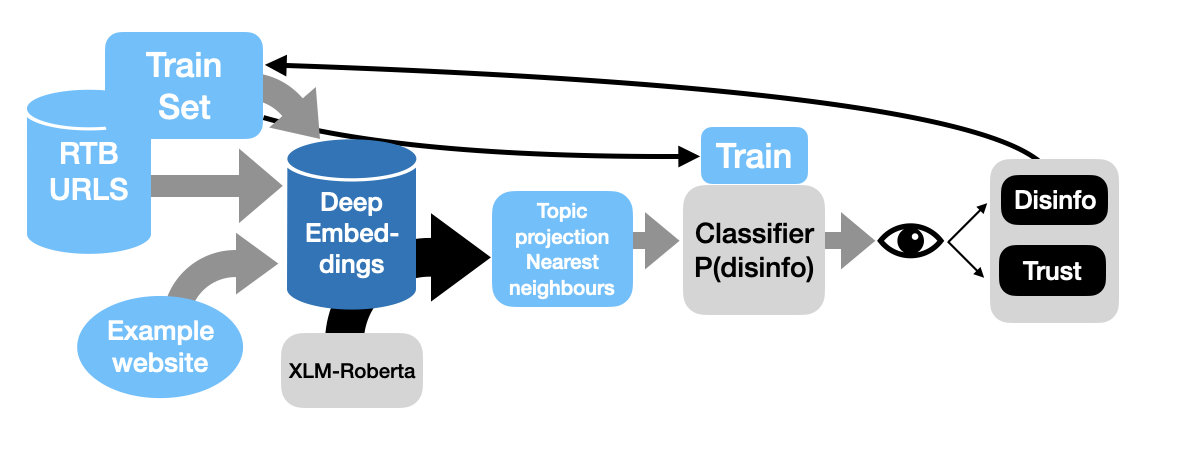}}
\caption{Disinformation prediction through content embeddings and human-in-the-loop approach.}
\label{model_diagram}
\end{center}
\vskip -0.3in
\end{figure}

Our main task is to rank domains with respect to their probability of containing disinformation on a certain topic of interest. Our solution has three steps: first, we filter a dataset of publisher pages, keeping only those related to the topic of interest; next, we apply a classifier to predict the probability of the content being disinformation; finally, we aggregate the page-level scores at domain-level, and a human reviewer manually inspects the most suspicious domains to confirm the label. In order to filter our dataset of pages on a particular topic, we rely on an embedding-based representation of web pages. The disinformation classifier is trained in a supervised manner. In order to rank the domains, we compute an average page score per domain. After human revision, the newly identified domains can be added to the training set and the model, refined (see Figure \ref{model_diagram}).

\subsection{Topic Detector}
The first step in our disinformation detection pipeline is filtering on a specific topic. We developed a zero-shot topic classifier that leverages similarity scoring between multilingual page embeddings. These are extracted from an XLM-RoBERTa model \cite{Conneau2019} that had previously been fine-tuned for an unrelated multi-label text classification task. The topic classifier is topic-agnostic and does not require labeled data for our topic classification task. 
We determine whether a page talks about the topic by applying a manually defined threshold on the cosine similarity between the topic embedding and the page embedding.

In this section, we explain how we built the embeddings (for a given topic and an input web page), how to select the minimum threshold on the similarity scores, and finally how these two pieces can be applied together to build the topic classifier.

\subsubsection{Embedding Prediction}
\label{embedding-prediction}
We use the base version of XLM-RoBERTa (henceforth XLM-R), a multilingual language model based on the Transformer architecture \cite{Vaswani2017}, to compute the embedding by running the model inference on a part of the text crawled from the web page (see Figure \ref{embedding_prediction_schema}).
We leverage a pre-trained XLM-R model that we have fine-tuned on a multi-label classification task for categorizing publisher web pages (see \ref{sec_dataset} for more information about the task and datasets). The quantized version\footnote{We used the dynamic quantization provided by PyTorch. This method, which represents the model weights and activations as integer rather than floating point, has proven to be very useful for running deep models at scale: the model inference time decreases significantly at the cost of a small drop in performance.} of this fine-tuned model is running at scale in our production environment. 
Since the model has been trained on annotated datasets with text extracted from web pages, it has learned some domain-specific knowledge of the content. 
The text is the concatenation of the content in the HTML tags \footnote{Tags are ordered by decreasing order of importance: title $>$ description $>$ h1 $>$ h2 $>$ \dots $>$ h6 $>$ p.}. This input is tokenized and truncated at 96 tokens.
We extract the page embedding by taking the hidden state of the \texttt{<cls>} token in the last layer of the encoder, which gives us an embedding of dimension 768.
We further reduce the dimension to 100 by Gaussian random projection, which does not impact the quality of the embedding (see section \ref{sec:eval_quality_embeddings}). 

\subsubsection{Topic Projection}
\label{sec:topic_classification}
To filter the pages on a particular topic, we start by selecting a few pages to serve as examples, getting their corresponding embeddings and computing the average to get the topic embedding. 
We then compute the similarity score of the page with respect to the topic via the cosine similarity between the page and topic embeddings. Finally, we rank all pages in an unlabeled dataset by decreasing similarity score and, through manual inspection, select a threshold (see Figure \ref{topic_classification_schema}; note that, due to the heterogeneity of the embedding space, this threshold is topic-specific, so this process must be repeated for each new topic). 
The manual inspection process starts with a list of eligible thresholds, which is used to create buckets of similarity scores. 
For each bucket, the user visualizes a sample of web pages that belong to it. The optimal threshold is selected based on the relevance of the web pages sampled in the corresponding bucket: the selected bucket is the farthest one from 1 such that the majority of the pages in the bucket are relevant to the topic. 

\subsection{Disinformation Classification Model}
Although topic filtering reduces the number of pages to be reviewed from billions to thousands, this is still a very high number to be reviewed by a human team within a reasonable time. To further narrow the search, we propose to train a disinformation classifier and rank the pages according to the classifier's output score. The reviewers can then select the top domains for inspection.

\subsubsection{Repurposed Embeddings for Disinformation Detection (REDD)}
\label{sec_redd}
In order to predict the probability of disinformation, we introduce the Repurposed Embeddings for Disinformation Detection (REDD) model. This model takes as input the page embeddings and feeds them to a simple three-layer nonlinear classifier with scaled exponential linear units (SELUs). We train the model for several epochs on the topic-filtered dataset using a binary cross-entropy loss function and predict the scores per page. 

Note that there are, in fact, many other options for the architecture or the classifier itself. Preliminary experimentation suggested that there is not much difference among these setups, since the classification task seems to be picked up easily, so we did not pursue this direction (other options would be to also compare against kNN, SVM, etc). In Section~\ref{sec:text-classifier} we compare the nonlinear REDD with a one layer MLP.

\subsubsection{Dataset}
\label{sec_dataset}
In order to train REDD, we use 60 domains already blocked for spreading disinformation, obtained from external and internal providers. Pages (URLs) from these domains serve as the disinformation examples and are associated with the positive label. On the other side, we manually select renowned media with a good reputation in fact-checking to get web pages with a negative label. The train dataset size is 17,473 samples and the test set size is 1,925.
The sets contain $53\%$ of positive (disinformation) examples. Both datasets have already been projected on a particular topic. The articles are in various languages, with some languages prevailing over others. This happens because disinformation in the topic of interest might be more prevalent in some languages than in others\footnote{This problem of modeling the sources rather than the content has already been encountered in the literature. See, e.g., \citealt{SanMartino2020}.}. This causes our dataset to be skewed by language, making the multilingual generalization challenging: there is a risk that the model might simply become a language detector, which we aim to prevent (see Section \ref{sec:text-classifier}).

\section{Experiments}
\label{sec:experiments}
We first evaluate the quality of fine-tuned embeddings, which will be used for the topic projection and as the input to the REDD model. We compare it with embeddings extracted from the general-purpose NLP model.
In the next experiment, we demonstrate the importance of the topic projection step by comparing against a model trained on a dataset that has not been filtered on the selected topic. Finally, we compare the performance of our REDD model, which takes embeddings as input, against a classifier trained directly on the text and discuss its impact on the multilingual task. 


\subsection{Repurposed Topic Embeddings}
\label{sec:eval_quality_embeddings}

In order to evaluate the quality of the embeddings, we employ a metric that measures how web pages with similar content are close in the embedding space. The goal of these experiments is to compare the performance of our embeddings with embeddings coming from pre-trained open-source models and also study the impact of dimensionality reduction on the quality of the embeddings.

The datasets for the embedding quality evaluation consist of web pages and their associated categories. These categories have been manually selected by humans for web pages in 13 languages and belong to the IAB content taxonomy\footnote{We use the v2.2 version of \url{https://iabtechlab.com/standards/content-taxonomy}.}. Details on the 13 datasets, their language and number of web pages are shown in Table \ref{targeting-segments-datasets-info}.

The \textbf{Probability of Same Categories} score (pSameCat) measures the average probability that the nearest neighbors of a web page share its categories. The metric assumes that web pages that have their embeddings close in the vector space should share related topics (represented by their web page categories). 
 In our case, for each web page, we search for the 5 nearest neighbors and compute the proportion that shares at least one category with the page. The metric for the dataset is the average of these individual scores.


For our 13 datasets, we evaluated the pSameCat scores of the embeddings extracted from four models. We also evaluate this metric for these embeddings after a dimension reduction to 100 dimensions using Gaussian random projection. 
The first model is our XLM-R model, fine-tuned for the multi-label classification task on our internal annotated data. The three other models are open-source pre-trained multilingual Transformer-based models: mBERT (multilingual BERT; \citealt{Devlin2018}), distilmBERT (a distilled version of mBERT; \citealt{Sanh2019}), and XLM-R (not fined-tuned on our data).


\begin{table*}[h]
\scriptsize
\caption{$\text{pSameCat} \times 100$, for datasets in several languages. Higher is better.}
\label{pSameCat-scores-info}
\centering
\vspace{3mm}
\begin{tabular}{p{0.15\linewidth}p{0.03\linewidth} ||p{0.03\linewidth}|p{0.03\linewidth}|p{0.03\linewidth}|p{0.03\linewidth}|p{0.03\linewidth}|p{0.03\linewidth}|p{0.03\linewidth}|p{0.03\linewidth}|p{0.03\linewidth}|p{0.03\linewidth}|p{0.03\linewidth}|p{0.03\linewidth}|p{0.03\linewidth}|p{0.03\linewidth}}
\hline
\textbf{Model embeddings} & \textbf{Dim} & \textbf{EN} & \textbf{FR} & \textbf{DE} & \textbf{JA} & \textbf{IT} & \textbf{ES} & \textbf{PT} & \textbf{TR} & \textbf{NL} & \textbf{AR} & \textbf{RU} & \textbf{KO} & \textbf{ZH} & \textbf{Avg}\\
\hline
pre-trained distilmBERT & 768 & 43.2 & 49.3 & 46.7 & 36.9 & 49.3 & 49.7 & 45.5 & 47.0 & 47.0 & 48.5 & 47.1 & 40.6 & 57.6 & 46.8 \\
pre-trained mBERT & 768 & 29.9 & 36.9 & 32.7 & 32.2 & 36.3 & 39.2 & 34.8 & 43.4 & 32.9 & 42.3 & 37.2 & 41.7 & 48.8 & 37.6\\
pre-trained XLM-R & 768 & 23.0 & 32.3 & 30.6 & 29.9 & 27.9 & 32.9 & 27.9 & 39.8 & 32.4 & 37.4 & 35.0 & 40.5 & 44.5 & 33.4\\
fine-tuned XLM-R & 768 & \textbf{75.4} & \textbf{75.6} & \textbf{75.2} & \textbf{59.8} & \textbf{70.9} & \textbf{73.7} & \textbf{69.8} & \textbf{57.0} & \textbf{74.1} & \textbf{61.4} & \textbf{73.0} & \textbf{67.0} & \textbf{75.9} & \textbf{69.9}\\

\hline
pre-trained distilmBERT & 100 & 36.6 & 43.8 & 41.2 & 33.0 & 44.4 & 44.5 & 40.2 & 44.3 & 42.4 & 44.6 & 41.5 & 36.0 & 53.5 & 42.0 \\
pre-trained mBERT & 100 & 23.6 & 29.6 & 26.1 & 25.8 & 28.3 & 31.8 & 26.7 & 39.6 & 26.9 & 37.8 & 28.7 & 35.8 & 40.8 & 30.9\\
pre-trained XLM-R & 100 & 16.6 & 26.4 & 22.7 & 23.3 & 22.4 & 27.7 & 22.6 & 33.9 & 26.1 & 32.9 & 27.5 & 34.6 & 36.2 & 27.1\\
fine-tuned XLM-R & 100 & \textbf{74.7} & \textbf{74.7} & \textbf{74.5} & \textbf{58.9} & \textbf{70.0} & \textbf{72.8} & \textbf{68.9} & \textbf{56.1} & \textbf{72.6} & \textbf{60.6} & \textbf{72.3} & \textbf{65.9} & \textbf{75.1} & \textbf{69.0}\\
\hline
\end{tabular}
\end{table*}

The pSameCat scores for all datasets and model embeddings are depicted in Table \ref{pSameCat-scores-info}. For both raw and dimension-reduced embeddings, the fine-tuned XLM-R model significantly outperforms all the pre-trained models across all languages.
This shows that fine-tuning moved web pages that share similar content and topics closer to each other in the embedding space.
Moreover, after we apply dimensionality reduction on these embeddings, their quality remains high with only a small drop of 1.3\% in average pSameCat score, a good signal compression.  
We chose therefore to use the dimension-reduced embeddings in our system.


\subsection{The Necessity of the Topic Projection Step}

As we mentioned before, our method consists of three steps: topic projection, prediction of the probability of disinformation, and human review. In order to justify the necessity of topic projection to build the disinformation ranking model, we trained an end-to-end model on the unfiltered dataset, i.e. on the dataset without the topic projection. We compare this setup against training the model on the topic-filtered dataset, described in Section~\ref{sec_dataset}. 
Without the topic projection, the model achieves an AUC-ROC on the (topic-projected) test set of 0.65 compared to 0.955 (see Table \ref{text-embeddings-table}). This shows that the topic filtering step is crucial to be able to capture the particular disinformation content.

We are aware that having to build a specific model per topic bears the disadvantage of not generalizing to other topics. However, this is not an issue for our particular application, since our production pipelines allow for such design, and the review process is intended to be iterative.

\subsection{Embeddings versus Text as the Model Input}
\label{sec:text-classifier}

\begin{table}[t]
\caption{Comparison of embeddings (REDD) versus text as input for the binary classifier.}
\label{text-embeddings-table}
\vskip 0.15in
\begin{center}
\resizebox{.48\textwidth}{!}{
\begin{tabular}{lrrrr}
\toprule
Model & AUC-ROC & Loss & Precision@50 & Precision@500\\
\midrule
REDD linear   & 0.868 & 0.405& 0.96 & 0.752\\ 
REDD non-linear   & \textbf{0.955} & \textbf{0.251}& \textbf{0.98} & \textbf{0.922}\\
REDD no topic filter   & 0.65 & 2.431 & 0.64   & 0.57     \\
Text fully trained & 1    & 0.493  & 1    & 1     \\
Text embeddings frozen & 1    & 0.675  & 0.98    & 0.998      \\
\bottomrule
\end{tabular}
}
\end{center}
\vskip -0.1in
\end{table}

\begin{table}[t]
\caption{Comparison of three models on language ID 24 versus the other languages.
}
\label{label-langid24-table1}
\vskip 0.15in
\begin{center}
\resizebox{.48\textwidth}{!}{
\begin{tabular}{cccccc}
\toprule
\multicolumn{2}{c}{} & \multicolumn{2}{c}{\textbf{Trustworthy}}& \multicolumn{2}{c}{\textbf{Disinformation}}\\ 
\multicolumn{1}{c}{}& \textit{}&  \multicolumn{1}{c}{\textit{24}}       & \textit{not 24} & \multicolumn{1}{c}{\textit{24}}       & \textit{not 24} \\ 
\toprule
\multicolumn{1}{l}{REDD}     & mean & \multicolumn{1}{c}{0.731}    & 0.115  & \multicolumn{1}{c}{0.834}    & 0.514  \\ 
\multicolumn{1}{c}{}& std  & \multicolumn{1}{c}{0.278}    & 0.196  & \multicolumn{1}{c}{0.194}    & 0.422  \\ \midrule
\multicolumn{1}{l}{Text XLM-R Fully Trained}     & mean & \multicolumn{1}{c}{{0.999}}    & 0.005  & \multicolumn{1}{c}{{0.999}}    & 0.333  \\ 
\multicolumn{1}{c}{} & std  & \multicolumn{1}{c}{5e-08} & 0.061  & \multicolumn{1}{c}{3e-08} & 0.516  \\ \midrule
\multicolumn{1}{l}{Text XLM-R Embeddings Frozen} & mean & \multicolumn{1}{c}{{0.355}}    & 0.277  & \multicolumn{1}{c}{{0.367}}   & 0.298  \\ 
\multicolumn{1}{c}{}& std  & \multicolumn{1}{c}{0.009}    & 0.011  & \multicolumn{1}{c}{0.01}     & 0.024  \\ 
\bottomrule
\end{tabular}
}
\end{center}
\vskip -0.1in
\end{table}

In this experiment, we compare training the classifier described in Section~\ref{sec_redd} on the raw text instead of the embeddings. The training and test datasets are described in Section~\ref{sec_dataset}: they contain the text of these articles and the fine-tuned embeddings from XLM-R.
The results are reported in Table~\ref{text-embeddings-table}. Surprisingly, the XLM-R model trained on textual content achieved an AUC-ROC of 1 on the test set. After examining the predictions, we found that the model learned to detect a particular language instead of any disinformation topic. That is: it predicts high scores if a page is in one particular language, in which the disinformation topic is more prevalent, while consistently assigning low scores to any page in any other language. We show the predicted scores per category and language in Table \ref{label-langid24-table1}.  This issue is expected for our disinformation task, and finding a workaround is not trivial\footnote{One could try translating the whole training set in a single language or ensure that it contains a broad mixture of languages.}. Note that our dataset has a strong language bias, so the evaluation of this issue is particularly hard: in the language where disinformation occurs most often, there are almost no negative examples.

\begin{figure}[ht]
\begin{center}
\caption{Distribution of predicted disinformation scores across various languages. The full-text model only learns to distinguish one language from others (orange and green), while the embedding-based model actually trains for the disinformation task.}
\centerline{\includegraphics[width=\columnwidth]{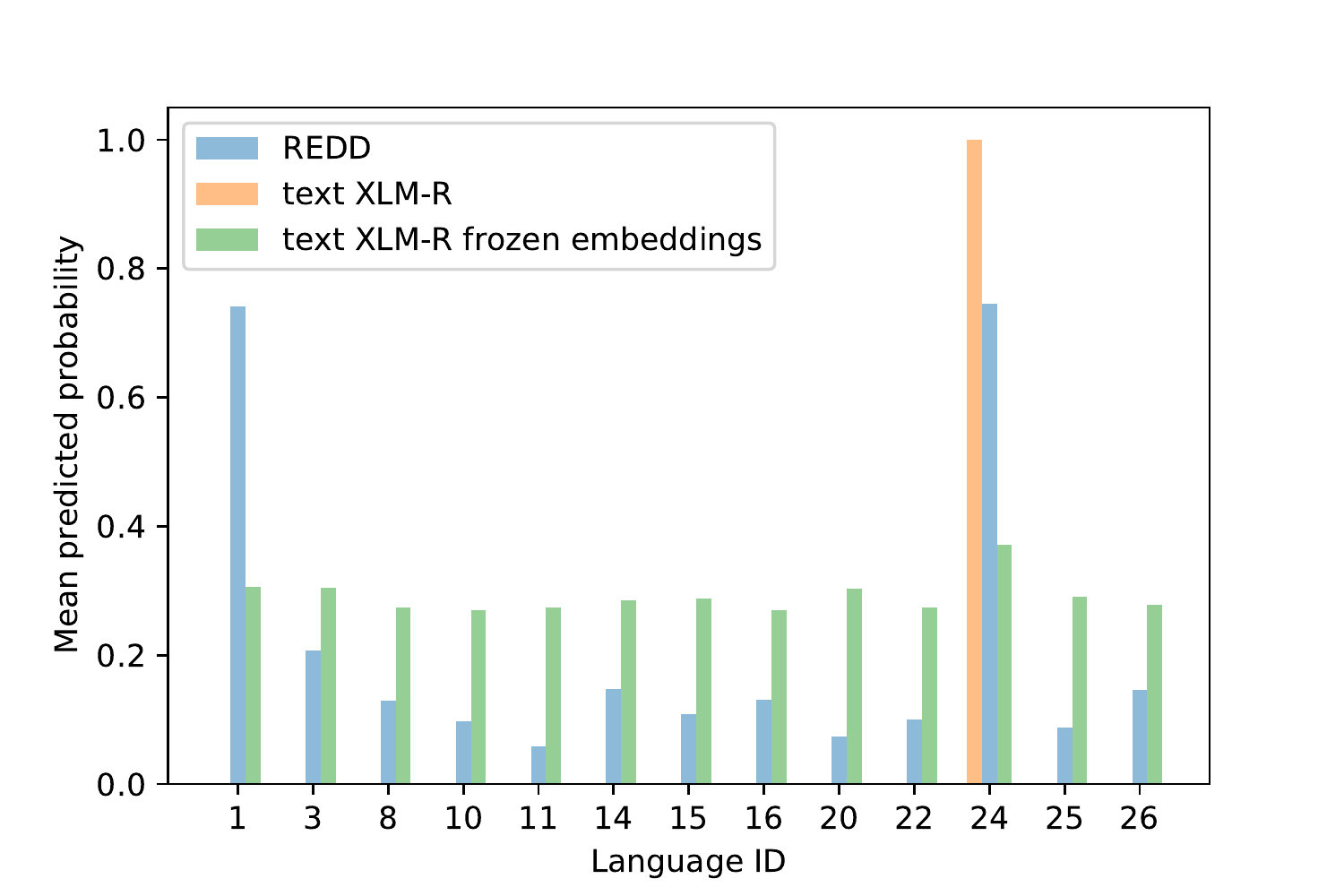}}
\label{fig:languages}
\end{center}
\vskip -0.2in
\end{figure}

However, for the REDD setup, where we take the fine-tuned embeddings as the input, the model focuses on the content instead of the language and is able to separate the disinformation content. We believe that the reason behind this is that the embeddings are already in a multilingual space, hence the training for disinformation prediction is language-agnostic. When using 100-dimensional (fixed) embeddings as the input to the nonlinear classifier, we obtain an AUC-ROC of 0.955, but the predicted probabilities are more equally distributed across the languages, see Table~\ref{label-langid24-table1} and Figure~\ref{fig:languages}. 

The main results are presented in Table~\ref{text-embeddings-table}, where we summarize the comparison of the following models: \textit{REDD linear}: a linear model taking the embeddings as input; \textit{REDD non-linear}: a model with three non-linear (SELU) layers, taking the embeddings as input; \textit{REDD no topic filter}: model trained on the unfiltered dataset and evaluated on the filtered test set; \textit{text fully trained}: XLM-R with a classification head, fine-tuned end-to-end on the text of the pages; and \textit{text embeddinds frozen}: XLM-R with a classification head, fine-tuned on the text of the pages, but with the embeddings kept frozen during the training.
We observe that the non-linear version of REDD outperforms the linear version and that the model with topic filtering overperforms the model without one. The seemingly good performance of the text-based model over the embedding-based model is actually due to the former only learning to detect one particular language.

\subsection{Human Review}
\begin{figure}[ht]
\begin{center}
\centerline{\includegraphics[scale=0.5]{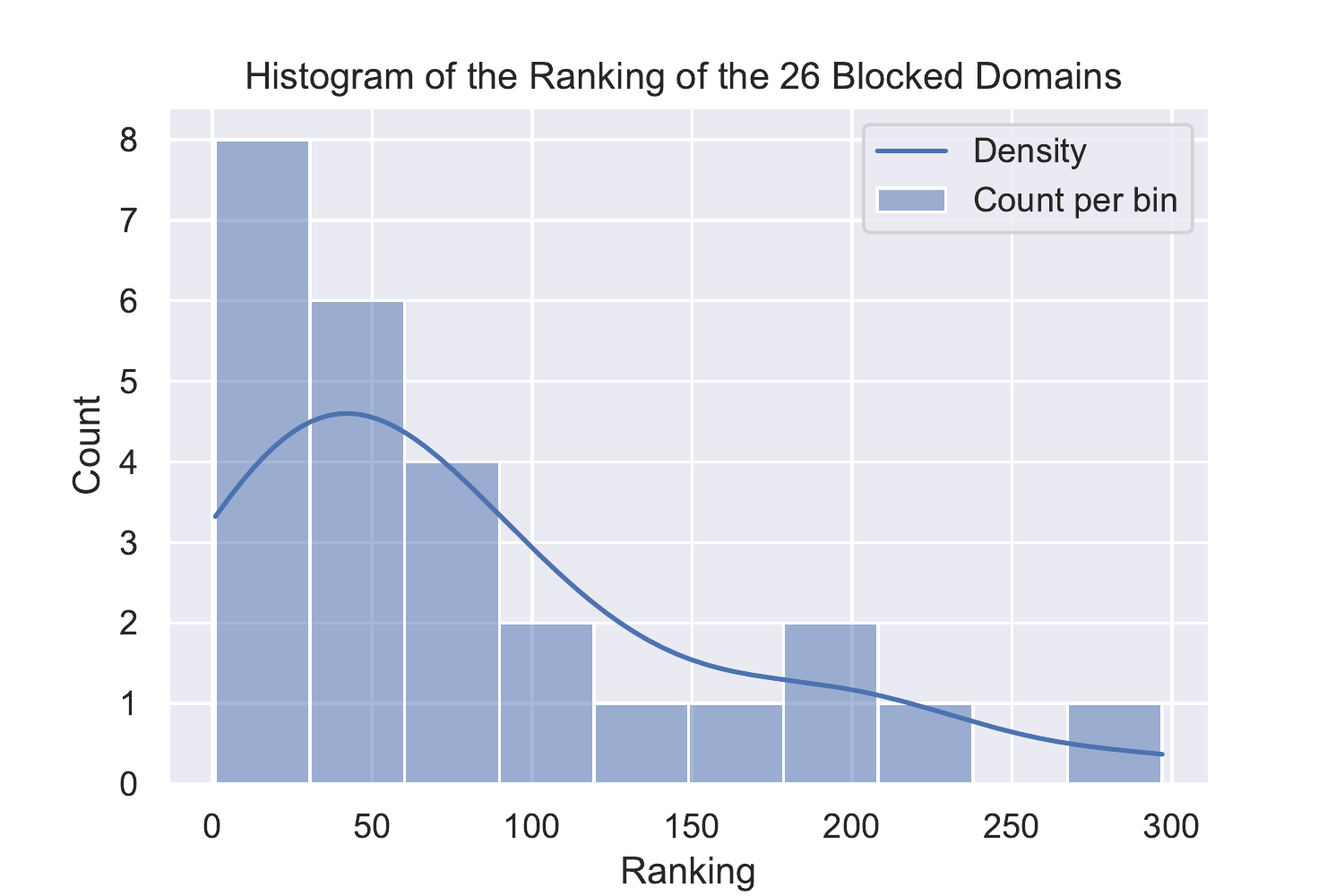}}
\caption{The distribution of the ranking position of the 26 identified disinformation domains among the 300 submitted for human review: the finally blocked domains are ranked higher by REDD.}
\label{ranking}
\end{center}
\vskip -0.3in
\end{figure}

The human review is done on the domain level: the domain score is computed as the average score of the pages in the domain. Ranking the domains by the highest scores helps to reduce the number of pages that need to be reviewed. We evaluated REDD on a set of articles prefiltered on the topic of interest. The predicted disinformation scores were aggregated to obtain a score per domain. This corresponded to 4000 ranked domains, out of which we sent the top 300 for human review. Out of these 300, 178 were flagged as suspected of spreading propaganda, and 26 were directly blocklisted from the publisher inventory\footnote{Our internal review process has several stages, where the whole content of the publisher is considered by the expert team.}. In Figure~\ref{ranking}, we show the distribution of the ranking of the blocked domains: these domains are more likely to be ranked higher by REDD among the top 300 reviewed domains. The evaluated precision at $k$ with $k=40$ equals $0.282$, significantly above the random baseline at $26/300=0.086$.

 Even though the number of blocked publishers might seem low, these were newly discovered domains, previously uncaught by external providers. This shows that the flexibility of our approach allows for adapting to industry-specific requirements. Finally, since our model works across many languages, we have identified disinformation domains on the same topic across various countries and languages.

\section{Discussion and Future Work}
In this work, we tackled the problem of disinformation identification by building REDD, a tool that red-flags domains that potentially spread disinformation. Even though the obtained probability predictions are not reliable for an automated decision, the tool allowed us to identify several domains not previously blocklisted. It can also support different languages and topics of interest, potentially preventing the spread of disinformation across different industries.

We have shown that filtering the training data on the topic of interest is necessary to obtain satisfactory results. Moreover, we demonstrated that leveraging fine-tuned embeddings helps the multilingual model focus on the disinformation task instead of on the language.

Our approach can be considered as the first step in improving publisher content analysis for digital advertisement in a multilingual setting. In the future, we can leverage orthogonal sources of information such as images, employ adversarial training (which has been suggested in the literature to improve performance: see \citealt{Wang2018}), or factorize a user-based disinformation graph to obtain embeddings and concatenate them with the content-based embeddings.

\section{Acknowledgements}
We would like to thank Béranger Dumont, Nicolas Pennequin, Thibault Becker, Kamila Jańczyk and François Zolezzi for their useful comments and contributions to this project.

\bibliography{bibliography}
\bibliographystyle{icml2022}

\newpage
\appendix
\twocolumn

\section{Model architectures}


\subsection{Three step reduction of number of pages for review}

\begin{figure}[ht]
\vskip 0.2in
\begin{center}
\centerline{\includegraphics[scale=0.5]{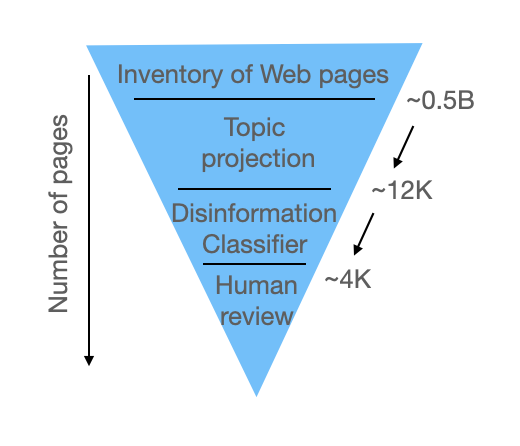}}
\caption{Topic projection and ranking of the pages reduces the number of pages that has to be reviewed by human reviewer.}
\label{topic_diagram}
\end{center}
\vskip -0.5in
\end{figure}

\vspace {0.5cm}
\subsection{Embedding prediction schema}

\begin{figure}[h!]
\begin{center}
\centerline{\includegraphics[width=\columnwidth]{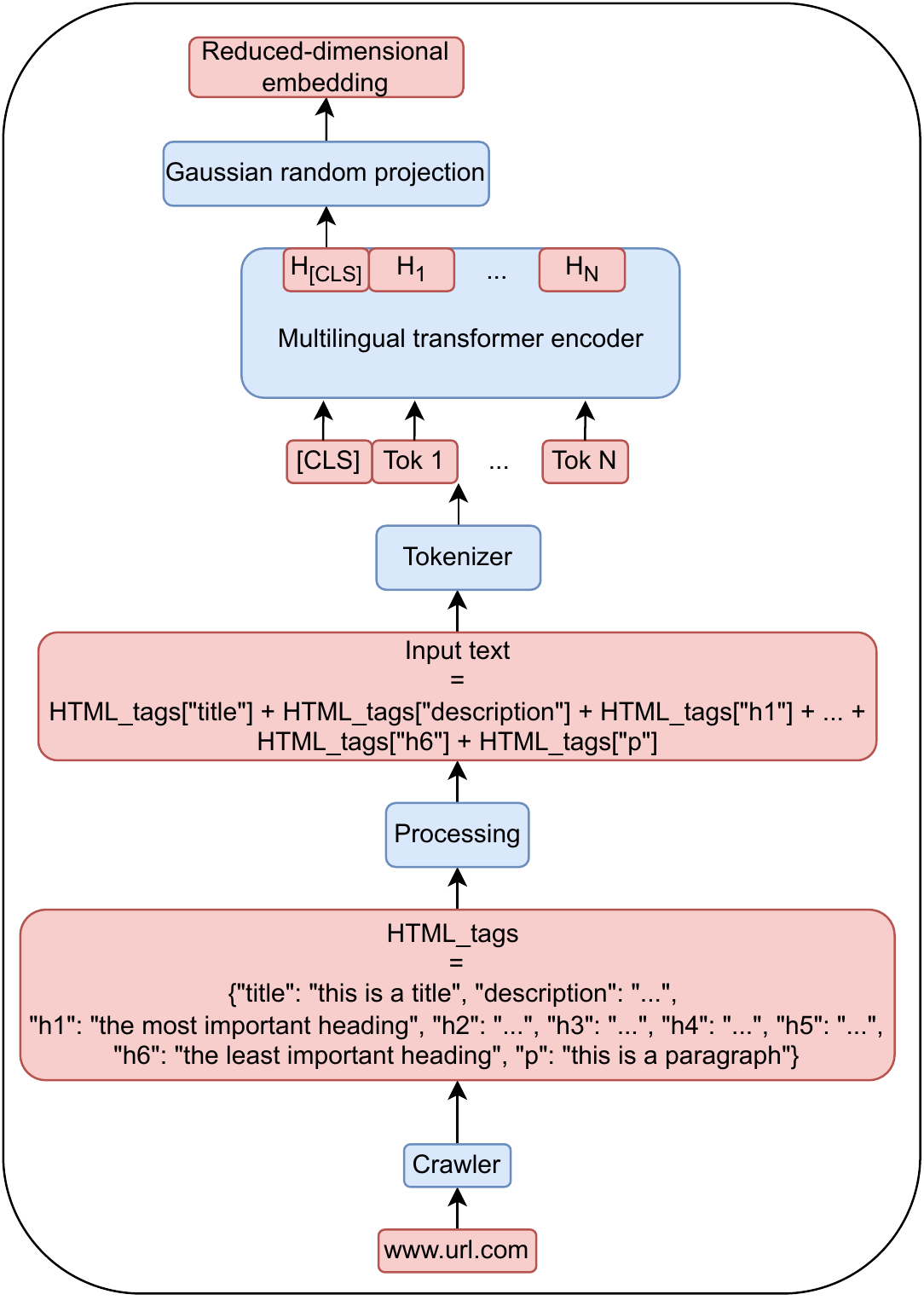}}
\caption{A detailed embedding prediction schema. }
\label{embedding_prediction_schema}
\end{center}
\end{figure}

\vspace {1cm}
\subsection{Topic classification schema}

\begin{figure}[ht]
\begin{center}
\centerline{\includegraphics[width=\columnwidth]{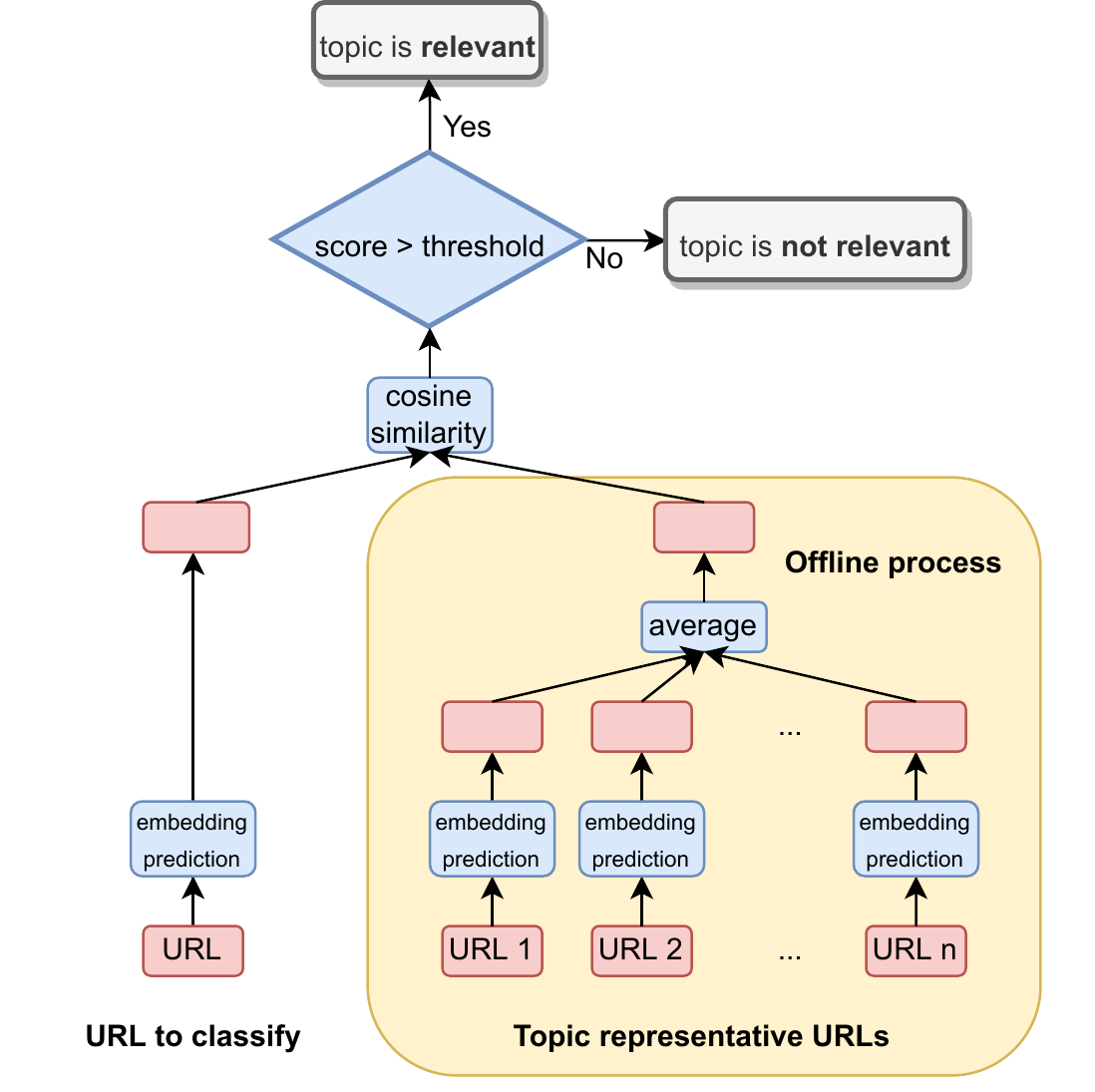}}
\caption{The topic classification schema. }
\label{topic_classification_schema}
\end{center}
\end{figure}

\newpage
\section{Additional details}

\begin{table}[h!]
\caption{Datasets with human-annotated web page categories. These datasets were used to evaluate the quality of the embeddings using the pSameCat metric. Some examples of categories are: News and Politics $>$ Politics, Technology and Computing $>$ Consumer Electronics $>$ Smartphones, and Travel $>$ Travel Type $>$ Air Travel.}
\label{targeting-segments-datasets-info}
\vskip 0.15in
\begin{center}
\begin{tabular}{clr}
\toprule
\multicolumn{2}{l}{\textbf{Language}} & \textbf{Size} \\
\midrule
EN & English & 53,290 \\
FR & French & 7,131 \\
DE & German & 5,276 \\
JA & Japanese & 7,772 \\
IT & Italian & 11,569 \\
ES & Spanish & 11,773 \\
PT & Portuguese & 9,081 \\
TR & Turkish & 11,409 \\
NL & Dutch & 6,921 \\
AR & Arabic & 6,302 \\
RU & Russian & 7,985 \\
KO & Korean & 7,304 \\
ZH & Chinese & 8,710 \\

\bottomrule
\end{tabular}
\end{center}
\vskip -0.1in
\end{table}

\end{document}